\documentclass[letterpaper,conference, 10pt]{ieeeconf}
\usepackage[T1]{fontenc}
\usepackage[utf8]{inputenc}
\usepackage{color}
\usepackage{units}
\usepackage{mathrsfs}
\usepackage{amsmath}
\usepackage{amssymb}

\makeatletter

\let\SF@@footnote\footnote
\def\footnote{\ifx\protect\@typeset@protect
    \expandafter\SF@@footnote
  \else
    \expandafter\SF@gobble@opt
  \fi
}
\expandafter\def\csname SF@gobble@opt \endcsname{\@ifnextchar[
  \SF@gobble@twobracket
  \@gobble
}
\edef\SF@gobble@opt{\noexpand\protect
  \expandafter\noexpand\csname SF@gobble@opt \endcsname}
\def\SF@gobble@twobracket[#1]#2{}
\providecommand{\tabularnewline}{\\}

\usepackage[printwatermark]{xwatermark}
\usepackage{xcolor}
\usepackage{graphicx}
\usepackage{tikz}
\usepackage{balance}

\makeatletter

\IEEEoverridecommandlockouts                              

\usepackage{amsmath} 
\usepackage{amssymb}  
\usepackage{import}

\usepackage{cite}
\usepackage{color}
\usepackage{graphicx}
\usepackage{subfigure}
\usepackage{cuted}
\makeatother

\makeatother

\begin{document}
\global\long\def\quat#1{\boldsymbol{#1}}

\global\long\def\dq#1{\underline{\boldsymbol{#1}}}

\global\long\def\hp{\mathbb{H}_{p}}

\global\long\def\dotmul#1#2{\langle#1,#2\rangle}

\global\long\def\partialfrac#1#2{\frac{\partial\left(#1\right)}{\partial#2}}

\global\long\def\totalderivative#1#2{\frac{d}{d#2}\left(#1\right)}

\global\long\def\mymatrix#1{\boldsymbol{#1}}

\global\long\def\vecthree#1{\operatorname{vec}_{3}#1}

\global\long\def\vecfour#1{\operatorname{vec}_{4}#1}

\global\long\def\haminuseight#1{\overset{-}{\mymatrix H}_{8}\left(#1\right)}

\global\long\def\hapluseight#1{\overset{+}{\mymatrix H}_{8}\left(#1\right)}

\global\long\def\haminus#1{\overset{-}{\mymatrix H}_{4}\left(#1\right)}

\global\long\def\haplus#1{\overset{+}{\mymatrix H}_{4}\left(#1\right)}

\global\long\def\norm#1{\left\Vert #1\right\Vert }

\global\long\def\abs#1{\left|#1\right|}

\global\long\def\conj#1{#1^{*}}

\global\long\def\veceight#1{\operatorname{vec}_{8}#1}

\global\long\def\myvec#1{\boldsymbol{#1}}

\global\long\def\imi{\hat{\imath}}

\global\long\def\imj{\hat{\jmath}}

\global\long\def\imk{\hat{k}}

\global\long\def\dual{\varepsilon}

\global\long\def\getp#1{\operatorname{\mathcal{P}}\left(#1\right)}

\global\long\def\getpdot#1{\operatorname{\dot{\mathcal{P}}}\left(#1\right)}

\global\long\def\getd#1{\operatorname{\mathcal{D}}\left(#1\right)}

\global\long\def\getddot#1{\operatorname{\dot{\mathcal{D}}}\left(#1\right)}

\global\long\def\real#1{\operatorname{\mathrm{Re}}\left(#1\right)}

\global\long\def\imag#1{\operatorname{\mathrm{Im}}\left(#1\right)}

\global\long\def\spin{\text{Spin}(3)}

\global\long\def\spinr{\text{Spin}(3){\ltimes}\mathbb{R}^{3}}

\global\long\def\distance#1#2#3{d_{#1,\mathrm{#2}}^{#3}}

\global\long\def\distancejacobian#1#2#3{\boldsymbol{J}_{#1,#2}^{#3}}

\global\long\def\distancegain#1#2#3{\eta_{#1,#2}^{#3}}

\global\long\def\distanceerror#1#2#3{\tilde{d}_{#1,#2}^{#3}}

\global\long\def\dotdistance#1#2#3{\dot{d}_{#1,#2}^{#3}}

\global\long\def\distanceorigin#1{d_{#1}}

\global\long\def\dotdistanceorigin#1{\dot{d}_{#1}}

\global\long\def\squaredistance#1#2#3{D_{#1,#2}^{#3}}

\global\long\def\dotsquaredistance#1#2#3{\dot{D}_{#1,#2}^{#3}}

\global\long\def\squaredistanceerror#1#2#3{\tilde{D}_{#1,#2}^{#3}}

\global\long\def\squaredistanceorigin#1{D_{#1}}

\global\long\def\dotsquaredistanceorigin#1{\dot{D}_{#1}}

\global\long\def\crossmatrix#1{\overline{\mymatrix S}\left(#1\right)}

\global\long\def\constraint#1#2#3{\mathcal{C}_{\mathrm{#1},\mathrm{#2}}^{\mathrm{#3}}}

\title{A Unified Framework for the Teleoperation of Surgical Robots in Constrained
Workspaces}

\author{Murilo~M.~Marinho\emph{,} Bruno~V.~Adorno, Kanako~Harada, Kyoichi~Deie,
Anton~Deguet, Peter~Kazanzides, \\Russell~H.~Taylor, and Mamoru~Mitsuishi\thanks{This
work was funded in part by the ImPACT Program of the Council for Science,
Technology and Innovation (Cabinet Office, Government of Japan), in
part by NSF grant 1637789, and in part by Johns Hopkins internal funds.}\thanks{Murilo
M. Marinho, Kanako Harada, and Mamoru Mitsuishi are with the Department
of Mechanical Engineering, the University of Tokyo, Tokyo, Japan.
\texttt{Emails:\{murilo, kanako, mamoru\}@nml.t.u-tokyo.ac.jp}. }\thanks{Murilo
M. Marinho's Ph.D. was supported by the Japanese Ministry of Education,
Culture, Sports, Science, and Technology (MEXT). His stay in Johns
Hopkins University was supported by the Graduate Program for Mechanical
Systems Innovation (GMSI), from the University of Tokyo}\thanks{Bruno
V. Adorno is with the Federal University of Minas Gerais, Brazil.
\texttt{Email: adorno@ufmg.br}. He has been supported by the Brazilian
agencies CAPES, CNPq (grants 424011/2016-6 and 303901/2018-7), FAPEMIG,
and by the INCT (National Institute of Science and Technology) under
the CNPq grant 465755/2014-3.}\thanks{Kyoichi~Deie is with the
Department of Pediatric Surgery, The University of Tokyo Hospital,
Tokyo, Japan.}\thanks{Anton Deguet, Peter Kazanzides, and Russell
H. Taylor are with the Department of Computer Science, Johns Hopkins
University, Baltimore, MD 21218 USA \texttt{Email:\{anton.deguet,pkaz,rht\}@jhu.edu}.}}
\maketitle
\begin{abstract}
In adult laparoscopy, robot-aided surgery is a reality in thousands
of operating rooms worldwide, owing to the increased dexterity provided
by the robotic tools. Many robots and robot control techniques have
been developed to aid in more challenging scenarios, such as pediatric
surgery and microsurgery. However, the prevalence of case-specific
solutions, particularly those focused on non-redundant robots, reduces
the reproducibility of the initial results in more challenging scenarios.
In this paper, we propose a general framework for the control of surgical
robotics in constrained workspaces under teleoperation, regardless
of the robot geometry. Our technique is divided into a slave-side
constrained optimization algorithm, which provides virtual fixtures,
and with Cartesian impedance on the master side to provide force feedback.
Experiments with two robotic systems, one redundant and one non-redundant,
show that smooth teleoperation can be achieved in adult laparoscopy
and infant surgery.
\end{abstract}

\section{Introduction}

The da Vinci Surgical System (Intuitive Surgical, Inc., Sunnyvale,
CA) has received considerable attention in the context of minimally
invasive surgery, which involves procedures performed through small
incisions. The robot is teleoperated: the surgeon generates motion
commands on the \emph{master side, }using a \emph{master} interface;
then, the commands are translated into motion by the \emph{slave }robot,
which interacts with the patient on the \emph{slave side}.

The success of the da Vinci in adult laparoscopy has led to attempts
to use it in surgical scenarios with workspaces more constrained than
those in the initial target applications, such as infant surgery \cite{Takazawa2018}
and paranasal sinuses and skull base surgery \cite{Schneider2013}.
However, these attempts have had limited success owing to the to large
diameter and length of the da Vinci's tools and its large operating-room
footprint. The fixed remote center-of-motion (RCM) is also a limitation.
Alternative designs try to compensate for some of those drawbacks
in adult laparoscopy \cite{lum2009raven,larocca2014new}.

Other robotic systems have been developed to provide assistance in
areas in which the da Vinci is hindered by its design. For instance,
robots have been developed for procedures in restricted workspaces
such as brain microsurgery \cite{mitsuishi2013master}, eye surgery
\cite{taylor1999steady}, endonasal surgery \cite{Burgner2014}, and
pediatric surgery \cite{Looi2013}. These robotic systems have several
designs, such as serial linkage, as in the da Vinci system and others
\cite{taylor1999steady,mitsuishi2013master}, parallel linkage \cite{Looi2013},
and flexible tubes \cite{Burgner2014,Leibrandt2017}. There are also
many control methodologies for the autonomous generation of constrained
motion using \emph{active constraints/virtual fixtures} \cite{funda1996constrained,kapoor2006constrained,li2007spatial,aghakhani2013task,pham2015analysis,Dahroug2017,marinho2018active,marinho2018dynamic}.

An in-depth survey on active constraints is presented by Bowyer \emph{et
al}. \cite{bowyer2014active}, who show that most of the research
in the field of virtual fixtures for teleoperated robots has focused
on impedance control on the master side, along with techniques such
as \emph{proxy and linkage simulation} and \emph{reference virtual
fixtures}. Impedance control on the master side has been successful
in pose\emph{}\footnote{\emph{Pose} stands for combined position and orientation.}
control of non-redundant robotic systems, such as the da Vinci, because
generating virtual fixtures on the master means the slave can be effectively
kept away from undesired interactions with the patient's anatomy \cite{vitrani2017applying}.

However, such techniques, if applied only on the master side, are
not suitable when the slave robot is redundant because, even if the
master's and the slave's end-effector poses are the same (with respect
to their own reference frames), the slave robot may have infinite
configurations in joint-space \cite{Siciliano2009}. Consequently,
some slave robot's links can have harmful interactions with the patient,
despite any feedback on the master. 

As an alternative to master-side techniques, slave-side techniques
have also been proposed, some of which use conventional control algorithms
based on the Jacobian pseudoinverse and nullspace projection \cite{Siciliano2009}
to generate an RCM \cite{aghakhani2013task,pham2015analysis,Dahroug2017}
or even more complex constrained workspaces \cite{Leibrandt2017}.
Nevertheless, these standard techniques struggle to deal with \emph{hard}
constraints\footnote{Hard constraints cannot be violated \cite{bowyer2014active}, in contrast
with \emph{soft} constraints \cite{kapoor2006constrained}, in which
small violations are allowed for short periods of time. } that are important in the medical field, such as joint and actuation
limits.

Considering hard limits, \emph{constrained optimization} \cite{funda1996constrained,kapoor2006constrained,li2007spatial,marinho2018active,marinho2018dynamic}
is a more suitable approach to designing motion control laws on the
slave side, because it naturally considers both inequality and equality
constraints, while taking into account all of the system's DOF. 

\section{Related works}

Initial approaches to constrained joint optimization in the generation
of virtual fixtures \cite{funda1996constrained,kapoor2006constrained,li2007spatial}
have been successful in providing constrained motion in complex scenarios,
but have had issues such as being ``computationally demanding and
inconsistent for some constraint and cost functions''\cite{bowyer2014active}.

The computational demand resulting from the use of quadratic positional
constraints and the difficulty of balancing virtual fixture and teleoperation
terms in the objective function is reported by Kapoor \emph{et al}.
\cite{kapoor2006constrained}. A follow-up work by Li \emph{et al}.\emph{
}\cite{li2007spatial} has been shown to be computationally more efficient,
as long as there is a single tool moving in the workspace , which
is not the case in most surgical scenarios. Kwok \emph{et al.} have
proposed \emph{ad hoc} techniques for snake robots \cite{kwok2013dimensionality}.
Lastly, several validation studies have focused on a single robotic
system in laparoscopic scenarios \cite{funda1996constrained,kapoor2006constrained,li2007spatial}
and in sinus surgery \cite{li2007spatial}, or on two robotic systems
that follow a predefined trajectory in the contexts of deep neurosurgery
\cite{marinho2018active} and transnasal surgery \cite{marinho2018dynamic}. 

A general framework for constrained motion control that does not depend
on specific robot designs can have several advantages. First, once
constraints are defined to achieve a desired behavior (e.g., avoiding
joint limits, preventing self-collisions and collisions with the workspace),
those same constraints can be applied to other type of robots to achieve
similar behavior. Second, the theoretical properties of the motion
controller (e.g., time response, closed-loop stability, computational
complexity) depend mostly on the framework, and a particular robotic
platform has little or no influence on the closed-loop behavior. Third,
researchers can focus on defining \emph{new} relevant constraints
for a particular robot design using a coherent theoretical framework,
instead of resorting to \emph{ad hoc }techniques. Thus, constrained
optimization allows for the the most generalizable solution, once
the aforementioned issues are solved. 

\subsection{Statement of contributions}

\textcolor{black}{In this paper, we propose a novel unified framework
for robot control under teleoperation, which is presented in Section~\ref{sec:Proposed-unified-framework}.
First, we tackle the issue of teleoperation in constrained optimization
approaches by proposing a teleoperation-oriented objective function,
without adding to it any virtual fixture terms, which facilitates
parameter tuning. Second, we combine the proposed objective function
with the vector field inequality method (VFI) \cite{marinho2018active,marinho2018dynamic}
to provide dynamic active constraints. Third, we add }\textcolor{black}{\emph{Cartesian
impedance}}\textcolor{black}{{} to our framework, effectively solving
the lack of haptic feedback of our earlier proposals \cite{marinho2018active,marinho2018dynamic}.}

These three contributions allow us to perform teleoperation in complex
scenarios, regardless of the to robot geometry. The generality of
the proposed unified framework is tested in two bi-manual experiments
using different robotic systems, as shown in Section~\ref{sec:Experiments}. 

\section{Mathematical Background}

The proposed unified framework for surgical robot teleoperation uses
quadratic programming for closed-loop inverse kinematics. To generate
dynamic virtual fixtures, geometrical primitives are modeled using
dual quaternion algebra, and linear constraints are added to the quadratic
program using the VFI method. The basics of quadratic programming
for closed-loop inverse kinematics, and the vector field inequalities
method are briefly explained in this section.

\subsection{Centralized quadratic programming for differential inverse kinematics
of multiple robots}

\textcolor{black}{Differential kinematics is the relation between
task-space velocities and joint-space velocities, in the general form
$\dot{\myvec x}=\mymatrix J\dot{\myvec q},$ in which $\myvec q\triangleq\myvec q\left(t\right)$
$\in$ $\mathbb{R}^{n}$ is the vector of manipulator joints' configurations,
$\myvec x\triangleq\myvec x\left(\myvec q\right)$ $\in$ $\mathbb{R}^{m}$
is the vector of $m$ task-space variables, and $\mymatrix J\triangleq\mymatrix J\left(\myvec q\right)$
$\in$ $\mathbb{R}^{m\times n}$ is a Jacobian matrix. The Jacobians
relating the robot's joint velocities to its end-effector's unit dual
quaternion pose ($\mymatrix J_{\dq x}$), rotation ($\mymatrix J_{r}$),
and translation ($\mymatrix J_{t}$) can be found using dual quaternion
algebra \cite{Adorno2011e}.}

\textcolor{black}{Suppose that $p$ robots should reach their own
independent task-space targets $\myvec x_{i,d}$ ($\dot{\myvec x}_{i,d}=\myvec 0$,
$\forall i,t$), for $i=1,\ldots,p$. Let each robot $R_{i}$ have
$n_{i}$ joints, joint velocity vector $\dot{\myvec q}_{i}$, task
Jacobian $\mymatrix J_{i}$, and task error $\tilde{\myvec x}_{i}=\myvec x_{i}-\myvec x_{i,d}$.
A suitable kinematic control law (assuming velocity inputs—i.e, $\myvec u\triangleq\dot{\myvec q}$)
with }\textcolor{black}{\emph{linear}}\textcolor{black}{{} constraints
is given by
\begin{align}
\myvec u=\underset{\dot{\myvec q}}{\arg\min\;} & \norm{\mymatrix J\dot{\myvec q}+\eta\tilde{\myvec x}}_{2}^{2}+\lambda\norm{\dot{\myvec q}}_{2}^{2}\label{eq:problem_quadratic_probots}\\
\text{subject to}\; & \mymatrix W\dot{\myvec q}\preceq\myvec w,\nonumber 
\end{align}
where
\begin{align*}
\mymatrix J & =\begin{bmatrix}\mymatrix J_{1} & \cdots & \mymatrix 0\\
\vdots & \ddots & \vdots\\
\mymatrix 0 & \cdots & \mymatrix J_{p}
\end{bmatrix}, & \myvec q & =\begin{bmatrix}\myvec q_{1}\\
\vdots\\
\myvec q_{p}
\end{bmatrix}, & \tilde{\myvec x} & =\begin{bmatrix}\tilde{\myvec x}_{1}\\
\vdots\\
\tilde{\myvec x}_{p}
\end{bmatrix},
\end{align*}
$\mymatrix W\triangleq\mymatrix W\left(\myvec g\right)$ $\in$ $\mathbb{R}^{r\times\sum n_{i}}$,
$\myvec w\triangleq\myvec w\left(\myvec g\right)$ $\in$ $\mathbb{R}^{r}$,
$\eta$ $\in$ $\left(0,\infty\right)$ is a proportional gain, and
$\mymatrix 0$ is a matrix of zeros with appropriate dimensions. The
damping factor $\lambda$ $\in$ $\left[0,\infty\right)$ provides
robustness to singularities \cite{cheng1994}}.

\subsection{Vector field inequalities in the generation of dynamic virtual fixtures}

The VFI method for dynamic elements \cite{marinho2018dynamic} first
requires a function $d\triangleq d(\myvec q,t)$ $\in$ $\mathbb{R}$
that encodes the (signed) distance between two geometric primitives.
Second, it requires a distance Jacobian and a residual relating the
time derivative of the distance function and the joints' velocities
in the general form
\begin{equation}
\dot{d}=\underbrace{\partialfrac{d(\myvec q,t)}{\myvec q}}_{\mymatrix J_{d}}\dot{\myvec q}+\zeta(t),\label{eq:distance_kinematics_general}
\end{equation}
where the residual $\zeta(t)=\dot{d}-\boldsymbol{J}_{d}\dot{\quat q}$
contains the distance dynamics unrelated to the joints' velocities.
The required distance function, distance Jacobians, and residuals
for all relevant primitives used in this paper are shown in \cite{marinho2018dynamic}.
\textcolor{black}{Lastly, the VFI method requires the definition of
a safe distance $d_{\text{safe}}\triangleq d_{\text{safe}}(t)$ $\in$
$\left[0,\infty\right)$ and a distance error $\tilde{d}\triangleq\tilde{d}(\myvec q,t)=d-d_{\text{safe}}$
to generate restricted zones or $\tilde{d}\triangleq d_{\text{safe}}-d$
to generate safe zones.} 

\textcolor{black}{With these definitions, and given $\eta_{d}$ $\in$
$\left[0,\infty\right)$, the signed distance dynamics for each pair
of primitives is constrained by
\begin{equation}
\dot{\tilde{d}}\geq-\eta_{d}\tilde{d}.\label{eq:velocity_damper}
\end{equation}
Constraint~\ref{eq:velocity_damper} assigns to each }primitive a
velocity constraint\textcolor{black}{{} that actively filters the robot
motion in the direction of the restricted zone boundary so that the
primitives do not collide. At most, each primitive will converge to
the boundary, and velocities tangential to the boundary itself are
unaffected.}

To use VFIs to generate restricted zones, we use the constraint
\begin{equation}
-\mymatrix J_{d}\dot{\myvec q}\leq\eta_{d}\tilde{d}+\zeta_{\text{safe}}\left(t\right),\label{eq:vfi_forbidden_zone}
\end{equation}
for $\zeta_{\text{safe}}\left(t\right)\triangleq\zeta\left(t\right)-\dot{d}_{\text{safe}}$.
Finally, safe zones are generated by using the constraint
\begin{equation}
\mymatrix J_{d}\dot{\myvec q}\leq\eta_{d}\tilde{d}-\zeta_{\text{safe}}\left(t\right).\label{eq:vfi_safe_zone}
\end{equation}

\section{Proposed unified framework\label{sec:Proposed-unified-framework}}

The proposed framework is divided into two parts, with a constrained
optimization algorithm that runs on the slave side and a Cartesian
impedance feedback that runs on the master side. Both are explained
in this section.

The technique proposed in this paper can be used to control any robotic
system, as long as the forward kinematics model and Jacobian are available.
Therefore, this includes serial-link, parallel-link, and even flexible
robots \cite{Leibrandt2017}.

\subsection{Slave side: Constrained optimization\label{subsec:constrained_optimization}}

Existing approaches to constrained optimization have terms in the
objective function for both trajectory tracking and virtual fixture
generation, which is a major source of parameter tuning difficulties
\cite{kapoor2006constrained} and inconsistencies in constraints and
cost functions \cite{bowyer2014active}. To prevent issues related
to having these mixed terms, the proposed technique includes only
those terms related to trajectory tracking in the objective function.

In the proposed framework, translation and rotation are represented
by quaternions. The quaternion set is $\mathbb{H}\triangleq\left\{ h_{1}+\imi h_{2}+\imj h_{3}+\imk h_{4}\,:\,h_{1},h_{2},h_{3},h_{4}\in\mathbb{R}\right\} $,
in which $\hat{\imath}^{2}=\hat{\jmath}^{2}=\hat{k}^{2}=\hat{\imath}\hat{\jmath}\hat{k}=-1$.
The conjugate of a quaternion $\quat h=h_{1}+\imi h_{2}+\imj h_{3}+\imk h_{4}$
is given by $\quat h^{*}=h_{1}-\left(\imi h_{2}+\imj h_{3}+\imk h_{4}\right)$
and $\vecfour{\quat h\triangleq\begin{bmatrix}h_{1} & h_{2} & h_{3} & h_{4}\end{bmatrix}^{T}}$.
Analogously, given a pure quaternion $\quat t=\imi x+\imj y+\imk z$,
we define $\vecthree{\quat t\triangleq\begin{bmatrix}x & y & z\end{bmatrix}^{T}}.$

Without loss of generality, suppose two identical slave robots are
controlled through teleoperation, each by an independent master interface
that generates a desired pose signal $\dq x_{i,d}$. In this paper,
we propose the following constrained optimization problem
\begin{alignat}{1}
\min_{\dot{\myvec q}}\  & \beta\mathscr{F}_{1}+\left(1-\beta\right)\mathscr{F}_{2}\label{eq:quadratic_problem_teleoperation}\\
\text{subject to}\  & \mymatrix W\dot{\myvec q}\preceq\myvec w,\nonumber 
\end{alignat}
where
\[
\mathscr{F}_{i}\triangleq\alpha f_{t,i}+\left(1-\alpha\right)f_{r,i}+f_{\Lambda,i},
\]
in which $f_{t,i}\triangleq\norm{\mymatrix J_{i,\quat t}\dot{\myvec q}_{i}+\eta\vecthree{\tilde{\myvec t}_{i}}}_{2}^{2}$,
$f_{r,i}\triangleq\norm{\mymatrix J_{i,\quat r}\dot{\myvec q}_{i}+\eta\vecfour{\tilde{\myvec r}_{i}}}_{2}^{2}$,
and $f_{\Lambda,i}\triangleq\norm{\mymatrix{\Lambda}\dot{\myvec q}_{i}}_{2}^{2}$
are the unweighted cost functions related to the end-effector translation,
end-effector rotation, and joint velocities of the $i$-th robot,
respectively; furthermore, each $i$-th robot has a vector of joint
velocities $\dot{\myvec q}_{i}$, a translation Jacobian (obtained
using $\vecthree{}$ instead of $\vecfour{}$ as in \cite{Adorno2011e,marinho2018active})
$\mymatrix J_{i,\quat t}$, a translation error $\tilde{\quat t}_{i}\triangleq\quat t_{i}-\quat t_{i,d}$,
a rotation Jacobian $\mymatrix J_{i,\quat r}$, and a switching rotational
error
\[
\tilde{\quat r}_{i}\triangleq\begin{cases}
\conj{\left(\myvec r_{i}\right)}\myvec r_{i,d}-1 & \text{if }\norm{\conj{\myvec r_{i}}\myvec r_{i,d}-1}_{2}<\norm{\conj{\myvec r_{i}}\myvec r_{i,d}+1}_{2}\\
\conj{\left(\myvec r_{i}\right)}\myvec r_{i,d}+1 & \text{otherwise},
\end{cases}
\]
based on the dual quaternion invariant error \cite{Figueredo2013},
where $\quat r_{i,d}$ and $\quat r_{i}$ are the desired and current
end-effector orientations, respectively. In addition, $\dot{\myvec q}=\begin{bmatrix}\dot{\myvec q}_{1}^{T} & \dot{\myvec q}_{2}^{T}\end{bmatrix}^{T}$
and $\mymatrix{\Lambda}\in\mathbb{R}^{n\times n}$ is a positive definite
damping matrix, usually diagonal. Lastly, $\alpha,\beta$ $\in$ $\left[0,1\right]$
are weights used to define the priorities between robots and between
the translation and the rotation. 

We use the linear constraints $\mymatrix W\dot{\myvec q}\preceq\myvec w$
to avoid joints limits \cite{cheng1994} and to generate active constraints
using the VFIs \cite{marinho2018dynamic}. Each parameter is explained
in more detail in the following subsections.

\subsubsection[The translation and rotation weight]{\label{subsec:translation-rotation-weight}The translation and rotation
weight, $\alpha$}

The weight $\alpha$ $\in$ $\left[0,1\right]$ is used to balance
translational and rotational gains. In our application, the translation
error is usually on a millimeter scale or lower. Therefore, the rotation
error may overtake the translation error, depending on the units used
to represent distance. Adding the weight $\alpha$ allows us to intuitively
set that balance without other modifications to the optimization problem.

\subsubsection[The robot prioritization weight]{\label{subsec:robot-prioritization-weight}The robot prioritization
weight, $\beta$}

The weight $\beta$ $\in$ $\left[0,1\right]$ is used to set a \emph{soft}
priority between robotic systems. To understand this parameter, first
note that if Problem~\ref{eq:quadratic_problem_teleoperation} has
a solution, the objective function will be optimized, \emph{given
that the linear constraints are satisfied}. This means that the linear
constraints prevent any collisions, even if this causes the trajectory
tracking error of a particular robot to increase. In such cases, the
parameter $\beta$ can be used to weight the priority between the
two robots. If $\beta>0.5$, then minimizing the trajectory tracking
error for robot 1 is favored over robot 2, effectively prioritizing
robot 1. The reverse is true for $\beta<0.5$. No explicit priority
is given if $\beta=0.5$.

\subsubsection[The joint weight matrix]{\label{subsec:joint-weight}The joint weight matrix, $\protect\mymatrix{\Lambda}$}

Whenever the robot is redundant and has a heterogeneous structure,
for instance a robotic manipulator with $n_{R}$ DOF attached to a
customized forceps with $n_{F}$ DOF, the damping matrix $\mymatrix{\Lambda}$
can be written in the form
\[
\mymatrix{\Lambda}\triangleq\begin{bmatrix}\mymatrix{\Lambda}_{R} & \mymatrix 0\\
\mymatrix 0 & \mymatrix{\Lambda}_{F}
\end{bmatrix},
\]
in which $\mymatrix{\Lambda}_{R}\in\mathbb{R}^{n_{R}\times n_{R}}$
and $\mymatrix{\Lambda}_{F}\in\mathbb{R}^{n_{F}\times n_{F}}$ are
matrices used to increase the relative weights of joints we wish to
have move less than others. For instance, given $\mymatrix{\Lambda}_{R}\triangleq\lambda_{R}\mymatrix I_{n_{R}}$
and $\mymatrix{\Lambda}_{F}\triangleq\lambda_{F}\mymatrix I_{n_{F}}$,
with $\lambda_{R},\lambda_{F}\in\left(0,\infty\right)$, we can favor
forceps motion over manipulator motion by setting $\lambda_{R}>\lambda_{F}$.

\subsubsection{\label{subsec:switching-controller}The switching unit quaternion
controller}

Because the group of unit quaternions double covers $\mathrm{SO}\left(3\right)$,
both $\quat r\in\mathbb{S}^{3}$ and $-\quat r$ represent the same
orientation, which causes the unwinding problem \cite{Kussaba2017}.
In practice, this problem results in undesired motions whenever a
continuous control law is employed. In order to see that, suppose
that the orientation error is given only by $\conj{\left(\myvec r_{i}\right)}\myvec r_{i,d}-1$;
if $\quat r_{i,d}=\quat r_{i}$, then the orientation error is equal
to $0$. However, if $\quat r_{i,d}=-\quat r_{i}$, then the orientation
error is equal to $-2$, although the current orientation is already
the desired one. In that case, the robot moves unnecessarily until
again reaching the new equilibrium point. A way to circumvent this
problem is to use discontinuous or hybrid control laws \cite{Kussaba2017},
which in our case is done by switching the error. This way, if $\conj{\left(\myvec r_{i}\right)}\myvec r_{i,d}$
is closer to $1$, the error is given by $\conj{\left(\myvec r_{i}\right)}\myvec r_{i,d}-1$;
conversely, if $\conj{\left(\myvec r_{i}\right)}\myvec r_{i,d}$ is
closer to $-1$, the error is given by $\conj{\left(\myvec r_{i}\right)}\myvec r_{i,d}+1$.

\subsection{Master side: Cartesian impedance\label{subsec:cartesian-impedance}}

In order to provide haptic feedback based on virtual fixtures, we
add a Cartesian force feedback on the master side that is proportional
to the current error on the slave side, in the form
\begin{align}
\quat{\Gamma}_{i,\text{master}} & \triangleq-\eta_{f}\tilde{\myvec t}_{i}^{\text{master}}-\eta_{V}\dot{\myvec t}_{i,\text{master}},\label{eq:cartesian_impedance}
\end{align}
for each master–slave pair, where $\quat{\Gamma}_{i,\text{master}}$
is the reflected force on the master side, $\eta_{f},\eta_{V}$ $\in$
$\left(0,\infty\right)$ are, respectively, stiffness and viscosity
parameters, $\tilde{\myvec t}_{i}^{\text{master}}$ is the translation
error of the slave, but seen from the point of view of the master,
and $\dot{\myvec t}_{i,\text{master}}$ is the linear velocity of
that master interface. This proportional force feedback with viscosity
allows the operator to ``feel'' any task-space directions in which
the robot has difficulty moving.

\section{Experiments\label{sec:Experiments}\protect\footnote{See accompanying video.}}

In order to evaluate the technique proposed in this paper, we first
present experiments to evaluate the effects of $\beta$ and the dynamic
active constraints using the da Vinci Research Kit (dVRK) \cite{kazanzides-chen-etal-icra-2014},
which is a research-friendly robotic system comprising the same master
and slave robotic systems of the da Vinci Surgical System. Second,
we present a peg transfer experiment to evaluate the proposed framework
in complex tasks. For this second experiment, a seven-DOF robot was
operated by a medical doctor.

The software implementation was the same for both systems, namely
Ubuntu 16.04 x64 running ROS Kinetic Kame.\footnote{http://wiki.ros.org/kinetic/Installation/Ubuntu}
Robot kinematics was implemented using the DQ Robotics\footnote{http://dqrobotics.sourceforge.net}
library, and constrained convex optimization was implemented using
IBM ILOG CPLEX Optimization Studio\footnote{https://www.ibm.com/bs-en/marketplace/ibm-ilog-cplex}
with Concert Technology.

\subsection{dVRK experiments}

\begin{figure}[h]
\centering

\def\svgwidth{1.0\columnwidth}

\scriptsize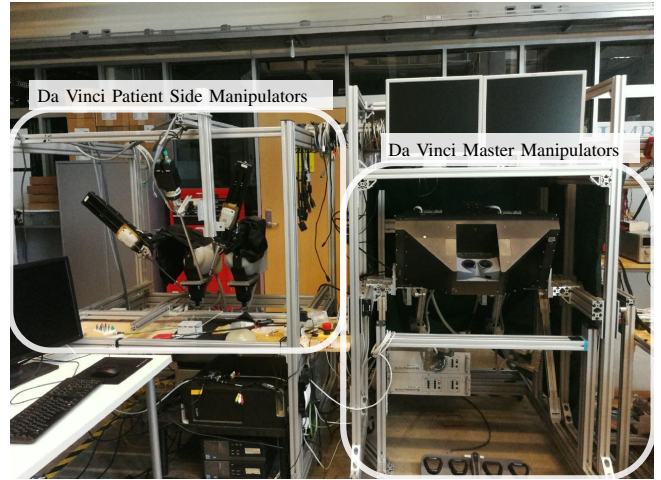

\caption{\label{fig:dvrk-setup} The dVRK experimental setup. Two slave arms
were commanded through two master arms.}
\end{figure}

The first set of experiments used the experimental setup shown in
Fig.~\ref{fig:dvrk-setup} and was devised to evaluate the effects
of a change in the prioritization weight $\beta$, while dynamic active
constraints to prevent collisions between shafts were enabled. Three
types of constraints were added: a shaft-to-shaft distance constraint,
to prevent collisions between tool shafts; a plane-to-point constraint,
to prevent collisions between the right tool and the peg transfer
board; and a joint limit constraint. All were implemented using VFIs
\cite{marinho2018dynamic}. 

The experiment involved manipulating a triangle on a peg transfer
board, which is the same peg transfer board used in the Fundamentals
of Laparoscopic Surgery (FLS) curriculum.\footnote{http://www.flsprogram.org}
For repeatability, before the task began, the right tool was positioned
on a central peg and the triangle was placed on the bottom-right peg
closest to the right tool. Only the left tool was allowed to move.
The right tool was commanded to stay in a constant pose throughout
the procedure.

The user had to pick and place the triangle in a clock-wise motion,
which required the triangle to be transferred between five pegs. Reaching
the four initial pegs should not induce any collisions between tools
and were useful to show whether the prioritization was cumbersome
outside of collision situations. The last target peg was the same
as the first to close the peg transfer circle. Reaching the last peg
required the left tool to push on the right tool's shaft. The behavior
of the system was evaluated under three different levels of prioritization,
as shown in Table~\ref{tab:dvrk-parameters}. The other parameters
are shown in Table~\ref{tab:all-parameters}.

\begin{table}[h]
\caption{\label{tab:dvrk-parameters} The priorities used in the dVRK teleoperation
experiments.}

\footnotesize{\centering%
\begin{tabular*}{1\columnwidth}{@{\extracolsep{\fill}}cccc}
\hline 
 & Same priority & Left tool higher priority & Left tool lower priority\tabularnewline
\hline 
$\beta$ & 0.5 & 0.99 & 0.01\tabularnewline
\hline 
\end{tabular*}}

\medskip{}

Suppose the tracking error of the tools should be 10~mm in order
to prevent a shaft–shaft collision. $\beta=0.5$ means that, in order
to prevent a collision, both arms' trajectory tracking errors are
increased by the same amount, therefore, 5~mm each. $\beta=0.99$
means that the left tool has a tracking error of $0.1$~mm and the
right tool an error of 9.9~mm. The reverse holds for $\beta=0.01$. 
\end{table}
\begin{table}[h]
\caption{\label{tab:all-parameters} Control parameters of Problem~\ref{eq:quadratic_problem_teleoperation}
used in each experiment. }

\footnotesize{\centering\textcolor{black}{}%
\begin{tabular*}{1\columnwidth}{@{\extracolsep{\fill}}cccccccccc}
\hline 
 & \textcolor{black}{$\alpha$} & \textcolor{black}{$\beta$} & \textcolor{black}{$\eta$} & \textcolor{black}{$\eta_{d}$} & \textcolor{black}{$\eta_{f}$} & \textcolor{black}{$\Lambda_{R}$} & \textcolor{black}{$\Lambda_{F}$} & \textcolor{black}{$\eta_{V}$} & \textcolor{black}{MS}\tabularnewline
\hline 
\textcolor{black}{dVRK} & \textcolor{black}{0.99} & \textcolor{black}{(Tab.~\ref{tab:dvrk-parameters})} & \textcolor{black}{1} & \textcolor{black}{1} & \textcolor{black}{350} & \textcolor{black}{$0.01$} & \textcolor{black}{$0.01$} & \textcolor{black}{10} & \textcolor{black}{$\unitfrac{1}{2}$}\tabularnewline
\textcolor{black}{Infant} & \textcolor{black}{0.99} & \textcolor{black}{0.5{*}} & \textcolor{black}{80} & \textcolor{black}{1} & \textcolor{black}{100} & \textcolor{black}{0.01} & \textcolor{black}{0.0} & \textcolor{black}{10} & \textcolor{black}{$\unitfrac{1}{3}$}\tabularnewline
\hline 
\end{tabular*}\textcolor{black}{}}

\medskip{}

{*}In the case of the infant experiment, there was no active constraint
relating both robots; therefore, we set $\beta=0.5$.

\textcolor{black}{$\eta$, $\eta_{d}$: proportional gain of the kinematic
controller and the VFI, respectively.}

\textcolor{black}{$\alpha$: translation error to orientation error
weight (Section~\ref{subsec:translation-rotation-weight}).}

\textcolor{black}{$\beta$: robot prioritization weight (Section~\ref{subsec:robot-prioritization-weight}).}

\textcolor{black}{$\Lambda_{R}$, $\Lambda_{F}$: Robot and forceps
joint gains, respectively (Section~\ref{subsec:joint-weight}). }

\textcolor{black}{$\eta_{F},\eta_{V}$: Cartesian impedance proportional
and viscosity gains, respectively (Section~\ref{subsec:cartesian-impedance}).}

\textcolor{black}{MS: Motion scaling. A motion scaling of X means
that a relative translation of the master was multiplied by X before
being sent to the slave.}
\end{table}

\subsubsection{Results and discussion}

\begin{figure}
\centering

\def\svgwidth{1.0\columnwidth}

\scriptsize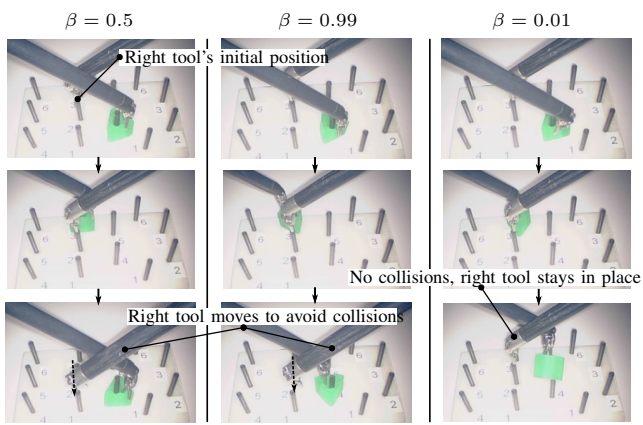

\caption{\label{fig:dvrk_peg_transfer} Snapshots of the dVRK experiments used
to evaluate the influence of parameter $\beta$. The first column
corresponds to $\beta=0.5$ (same priority), the second corresponds
to $\beta=0.99$ (left tool with higher priority), and the third column
corresponds to $\beta=0.01$ (left tool with lower priority).}
\end{figure}

\begin{figure}
\centering

\def\svgwidth{1.0\columnwidth}

\scriptsize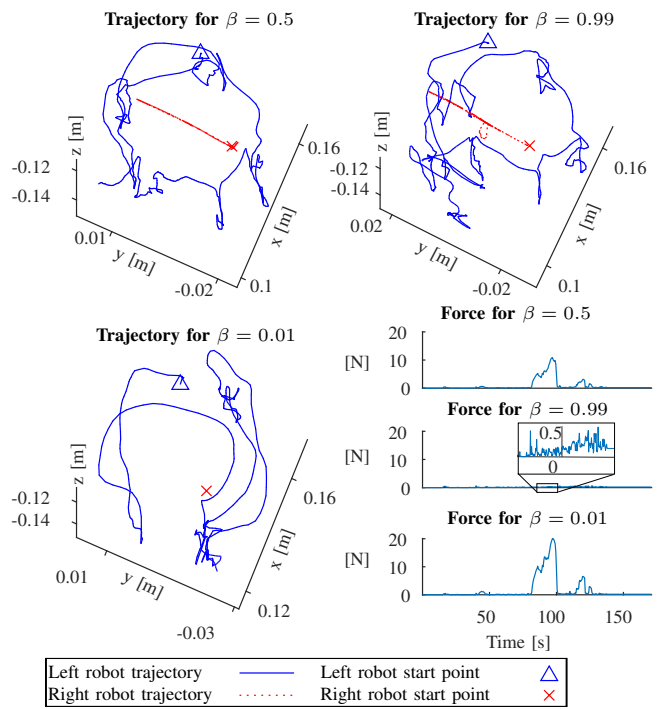

\caption{\label{fig:dvrktrajectoryandforce} Force feedback, based on virtual
fixtures, for the controlled tool and trajectories of both tools in
all three trials. The trajectories of the left and right tools are
shown in \emph{blue }and \emph{dotted red}, respectively.}
\end{figure}

\textcolor{black}{Snapshots of the peg transfer task using the dVRK
are shown in Fig.~\ref{fig:dvrk_peg_transfer}, for each of the three
experimental cases. Complete footage of each experiment is shown in
the accompanying video. Trajectory and force data for all cases are
shown in Fig.~\ref{fig:dvrktrajectoryandforce}.}

\textcolor{black}{In the first case ($\beta=0.5$), the left tool
could reach all pegs, as required by the task. The right tool autonomously
evaded the left tool whenever the left tool was commanded to a region
that would cause a collision. Although the positioning of the triangle
on the last peg was possible, it required considerable force from
the operator to push the right tool, which peaked at about 10N.}

\textcolor{black}{In the second case ($\beta=0.99$), the left tool
could reach all pegs, as required by the task. The force feedback
on the left tool was weak and barely distinguishable from the viscosity-induced
feedback; therefore, the left tool could even place the triangle on
the peg over which the right tool was initially located.}

\textcolor{black}{Finally, in the last case ($\beta=0.01$), the left
tool was not able to reach all pegs in the prescribed order. The user
could feel a strong force feedback whenever forcing the left tool
against the right tool. Even with considerable force from the operator,
20N, the right tool did not move away.}

\textcolor{black}{These results show that the parameter $\beta$ can
be used to prioritize tools in an intuitive manner. How to effectively
use this in a surgical task is left to future work.}

\subsection{Infant peg transfer experiments}

\begin{figure}
\centering

\def\svgwidth{1.0\columnwidth}

\scriptsize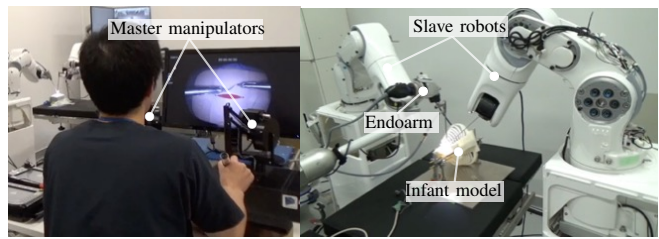

\caption{\label{fig:infant-peg-transfer} The master–slave configuration used
in the peg transfer infant experiments. The medical doctor commanded
two haptic interfaces and the robot automatically generated constraints
to avoid dangerous collisions with the infant model. }
\end{figure}

In this task, our target was to determine whether a medical doctor
could perform a difficult task under teleoperation in a constrained
workspace. Therefore, an expert in manual laparoscopic pediatric surgery
was invited to participate in this preliminary experiment.

The constraints in infant surgery are considerably more complex than
those in adult laparoscopy, and the da Vinci was shown to be inadequate
for this type of surgery \cite{Takazawa2018}. In this context, we
employed a surgical system that is being developed in parallel to
this work. 

Three types of constraints are required in infant surgery. First,
medical doctors use the compliance of the infant's skin to increase
the reachable workspace. This compliance can be considered in our
framework by generating an entry-sphere (shaft-to-point distance with
safe distance larger than zero), rather than using an entry point.
Second, the tool might move outside of the camera's field-of-view
owing to the small size of the workspace. Even though this situation
is common in manual surgery, because medical doctors rely on their
spatial perception of their bodies to locate tools, such out-of-bounds
motion is highly undesirable in robot-aided surgery owing to safety
concerns. In this context, a safety cuboid constraint was added for
each individual robotic arm. Lastly, joint limits were also considered.

As in the FLS curriculum, the medical doctor was asked to transfer
the triangles from one side of the peg transfer board to the other. 

\subsubsection{Results and discussion}

\begin{figure}
\centering

\def\svgwidth{1.0\columnwidth}

\scriptsize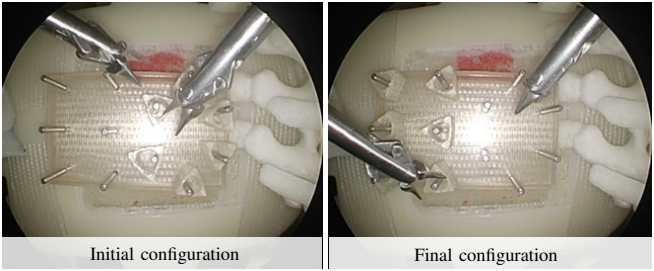

\caption{\label{fig:infant-peg-transfer-1} Snapshots of the initial and final
states of the peg transfer experiment with the medical doctor.}
\end{figure}
\begin{figure}
\centering

\def\svgwidth{1.0\columnwidth}

\small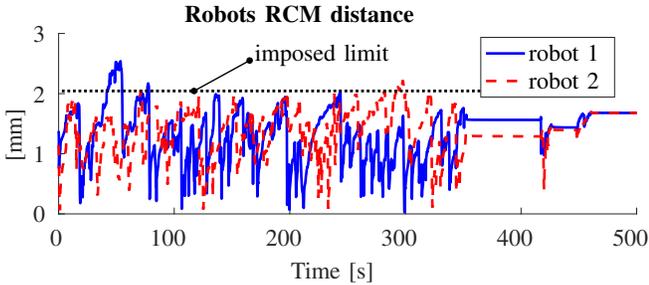

\caption{\label{fig:infant-peg-transfer-2-1} Distance of each robot's tool
shaft to the center of their respective entry-sphere.}
\end{figure}

\begin{figure}
\centering

\def\svgwidth{1.0\columnwidth}

\small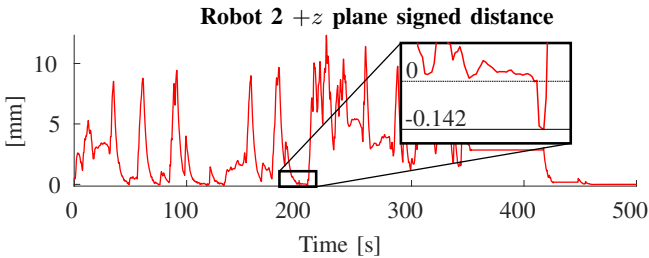

\caption{\label{fig:infant-peg-transfer-2-1-1} Distance of the second robot's
tool tip to the upper wall of the workspace cuboid, in which the constraint
was slightly violated. No effective violation was measured in the
other walls.}
\end{figure}

The medical doctor participated in three trials, one of which is illustrated
in Fig.~\ref{fig:infant-peg-transfer-1}. With very little experience
using the proposed system, the medical doctor was able to perform
a full peg transfer experiment in about 7~min. Overall, the medical
doctor gave a high evaluation of the robotic system usability. 

Qualitatively, after inspecting a video recording of the robot motion
during the peg transfer experiment, it was visible that the entry-sphere
constraint was properly maintained. There were no rib dislocations
and no model motion, which happened when using the da Vinci \cite{Takazawa2018}.

Quantitatively, the tool shaft distance to the entry-sphere center
is shown in Fig.~\ref{fig:infant-peg-transfer-2-1}, as measured
from the robot's encoders. The maximum distances between each robot
shaft and the center of its entry-sphere were 2.54~mm and 2.41~mm,
respectively. This means there was a maximum constraint violation
of 0.5~mm. Understanding the source of this constraint violation
is a topic of ongoing research. The culprit is thought to be the discrete
time implementation of Problem~\ref{eq:quadratic_problem_teleoperation}.

Another important set of constraints was the planar constraints making
up the cuboid workspace. Among the 12 plane constraints, the maximum
constraint violation corresponded to the plane that impeded the right
robot's tool tip from being retracted from the model, as shown in
Fig.~\ref{fig:infant-peg-transfer-2-1-1}. The magnitude of the violation
was 0.142~mm, which is of a similar magnitude to the constraint violation
of the entry-sphere. Other planes showed negligible constraint violations
of under 0.1~mm. Because the right robot tool tip was kept at the
border of that plane during most of the experiment, this indicates
why a higher violation of that plane was observed. 

These results show that a complex task, with several active constraints,
can be performed smoothly under teleoperation by a medical doctor
using the proposed framework. How well the framework can operate in
still more complex scenarios, including flexible tools, is a topic
of ongoing research.

\section{Conclusions and future work}

\textcolor{black}{In this paper, a novel unified framework for robot
control under teleoperation was proposed. The method can be used to
provide smooth teleoperation, regardless of the robot geometry and
under workspace constraints. On the slave side, a constrained optimization
algorithm provides virtual fixtures for collision avoidance and the
avoidance of joint limits. On the master side, a Cartesian impedance
algorithm allows the user to ``feel'' directions in which the robot
has difficulty moving. }The proposed framework is evaluated in two
scenarios, with different robot geometries. First, we demonstrate
a shaft–shaft collision avoidance with tool prioritization under teleoperation
using the dVRK. Second, we show a peg transfer experiment performed
by a medical doctor using a redundant robot system in an infant surgery
scenario.

In future works, we plan to test the performance of the framework
in the teleoperation of flexible robots.

\balance

\bibliographystyle{ieeetr}
\bibliography{bib/ral}
 
\end{document}